\documentclass[conference]{IEEEtran}
\IEEEoverridecommandlockouts

\usepackage{cite}
\usepackage{amsmath,amssymb,amsfonts}
\usepackage{algorithmic}
\usepackage{graphicx}
\usepackage{hyperref}
\usepackage{textcomp}
\usepackage{xcolor}
\def\BibTeX{{\rm B\kern-.05em{\sc i\kern-.025em b}\kern-.08em
    T\kern-.1667em\lower.7ex\hbox{E}\kern-.125emX}}
\begin{document}

\title{ReL-SAR: Representation Learning for Skeleton Action Recognition with Convolutional Transformers and BYOL\\

}

\author{\IEEEauthorblockN{Safwen Naimi}
\IEEEauthorblockA{\textit{Data Science Laboratory} \\
\textit{University of Quebec (TÉLUQ)}\\
Montréal, Canada} \\
\and
\IEEEauthorblockN{Wassim Bouachir}
\IEEEauthorblockA{\textit{Data Science Laboratory} \\
\textit{University of Quebec (TÉLUQ)}\\
Montréal, Canada} \\
\and
\IEEEauthorblockN{Guillaume-Alexandre Bilodeau}
\IEEEauthorblockA{\textit{LITIV lab.} \\
\textit{Polytechnique Montréal}\\
Montréal, Canada}\\
}

\maketitle

\begin{abstract}
To extract robust and generalizable skeleton action recognition features, large amounts of well-curated data are typically required, which is a challenging task hindered by annotation and computation costs. Therefore, unsupervised representation learning is of prime importance to leverage unlabeled skeleton data. In this work, we investigate unsupervised representation learning for skeleton action recognition. 
For this purpose, we designed a lightweight convolutional transformer framework, named ReL-SAR, exploiting the complementarity of convolutional and attention layers for jointly modeling spatial and temporal cues in skeleton sequences. We also use a Selection-Permutation strategy for skeleton joints to ensure more informative descriptions from skeletal data. Finally, we capitalize on Bootstrap Your Own Latent (BYOL) to learn robust representations from unlabeled skeleton sequence data. We achieved very competitive results on limited-size datasets: MCAD, IXMAS, JHMDB, and NW-UCLA, showing the effectiveness of our proposed method against state-of-the-art methods in terms of both performance and computational efficiency. To ensure reproducibility and reusability, the source code including all implementation parameters is provided at 
\href{https://github.com/SafwenNaimi/Representation-Learning-for-Skeleton-Action-Recognition-with-Convolutional-Transformers-and-BYOL}{https://github.com/SafwenNaimi}.
\end{abstract}


\section{Introduction}
Human activity recognition (HAR) is used in areas like video surveillance, sports analysis, robotics, and health monitoring. It aims to analyze videos containing humans to identify the performed actions. Despite significant progress, robustly recognizing a wide range of human activities remains difficult due to various factors present in realistic environments, such as large variations in how the same action can be performed under different viewpoints and by different persons, and the recording conditions that may not be favorable due to occlusions. 

Training HAR methods in a fully supervised manner requires large datasets with accurate annotations, which are time-consuming and costly to prepare in some crowded scenarios. To overcome this issue, self-supervised methods have been proposed to learn from data without the need for labels \cite{Zheng2018UnsupervisedRL,Su2019PREDICTC,Lin2020MS2LMS,Rao2020AugmentedSB,Grill2020BootstrapYO}. Those pre-trained representations generated from unlabeled data can then be fine-tuned by training on a smaller labeled dataset to create an efficient model to solve a specific task. The first approaches have been focusing on learning representations by solving pretext tasks \cite{Zheng2018UnsupervisedRL,Su2019PREDICTC}, while more recent works adopt a contrastive learning framework \cite{Lin2020MS2LMS,Rao2020AugmentedSB}. The goal of contrastive representation learning is to create an embedding space where comparable sample pairs are close together and dissimilar sample pairs are far away. Bootstrap Your Own Latent (BYOL) \cite{Grill2020BootstrapYO} follows this contrastive objective and achieves very good results in computer vision tasks without using negative pairs, by relying on two neural networks that interact and learn from each other.

By capitalizing on self-supervised learning, we propose a novel human activity recognition method addressing the limited data availability problem.
The proposed method is based on human skeletons to mitigate the impact of visual appearance variation. We first detect humans in the scenes and estimate their poses with skeletons over temporal windows. This provides a skeletal action representation focused on discriminative posture and gesture cues, rather than appearance details, like clothing. To capture the most salient features of these skeletons, a Selection-Permutation strategy is applied to the skeleton joints. We then introduce a lightweight convolutional transformer-based model to capture the relationships between skeleton joints, both spatially and temporally, which we called ReL-SAR (\emph{Re}presentation \emph{L}earning for \emph{S}keleton \emph{A}ction \emph{R}ecognition). The convolutional layers specialize in extracting local spatial features to discern the spatial hierarchies within human poses and ensure robustness to variability in appearance thanks to their spatial inductive bias. Then, the transformer component allows to model temporal dependencies and dynamic evolution of poses, leveraging the complementary strengths of transformer and convolutional layers. To overcome the need for large amounts of data, ReL-SAR is pre-trained with BYOL.
We show that representations learned with BYOL can effectively boost the performance.

The main contributions of this paper are as follows:
\textbf{1)} We propose a lightweight convolutional transformer-based model that we call ReL-SAR, to learn efficient representations for skeleton-based action recognition, by leveraging the benefits of both convolutional layers and the attention of transformers; 
\textbf{2)} We demonstrate that our model, using BYOL, can learn relevant features from unlabeled skeleton sequences and ensure robust low-level representations compared to those obtained through fully supervised learning, despite the absence of explicit labels;
\textbf{3)} We show that by removing irrelevant joints and ensuring a meaningful arrangement of joints within the input sequence, our Selection-Permutation strategy further enhances the performance of ReL-SAR.

\section{Related Work}

Skeleton action recognition uses skeleton data, which is robust to changes in appearance or environment, to analyze human actions by focusing on joint movements. Early research, such as Fernando et al. \cite{7299176}, utilized handcrafted features and hidden Markov models, while subsequent studies like Du et al. \cite{7298714} adopted LSTM-based recurrent neural networks to capture temporal dynamics of movements. However, these methods often overlooked spatial relationships among joints. More recent approaches, such as those by Liu et al. \cite{Liu2017EnhancedSV} and Yang et al. \cite{Yang2019MakeSA}, have incorporated both pose and motion information using two-stream RNN architectures and encoded pose images, respectively. Choutas et al. \cite{Choutas2018PoTionPM} developed PoTion, a method that aggregates joint heatmaps colored by relative time into a fixed-dimension representation for CNN-based action classification. Despite the spatial focus of CNNs, their limited capacity in modeling long-range temporal dependencies led to the exploration of Transformers, which excel in managing long-range dependencies through self-attention mechanisms, as noted by Dosovitskiy et al. \cite{dosovitskiy2021an}. This has motivated the integration of CNNs and Transformers to leverage the strengths of both models in recognizing human actions.
Self-supervised learning has evolved significantly in the field of action representation from skeleton sequences. Initial approaches focused on pretext tasks, with Zheng et al. \cite{Zheng2018UnsupervisedRL} utilizing a recurrent encoder-decoder GAN for learning representations, and Kundu et al. \cite{Kundu2018UnsupervisedFL} employing a variational autoencoder to capture the human pose manifold, subsequently mapping actions as trajectories within this space. Transitioning to contrastive learning, recent efforts such as Lin et al. \cite{Lin2020MS2LMS} integrated it into a multi-task learning framework (MS²L), aiming for more general representations. Thoker et al. \cite{Thoker2021SkeletonContrastive3A} explored cross-contrastive learning between graph-structured and sequence-structured data to retain both modality-invariant and specific features. Li et al. \cite{Li20213DHA} developed SkeletonCLR and CrosSCLR, leveraging a momentum contrast framework to enhance skeleton sequence representations. These methods traditionally use positive and negative sample pairs to train networks for distinguishing relevant pairings. However, the BYOL approach in ReL-SAR focuses only on positive pairs, which offers benefits like improved stability, better memory efficiency, and superior representation quality without the need for designing negative pairs.

\section{Proposed Method}

We motivate and explain our proposed ReL-SAR model in this section, covering pose extraction, our Selection-Permutation strategy, details of our convolutional transformer architecture, and the application of BYOL for representation learning.

\subsection{Human Detection and Pose Estimation}

\label{3.1}
Humans are first detected to establish regions of interest for subsequent processing. We use YOLOv5x  \cite{https://doi.org/10.5281/zenodo.3908559} for this purpose. We then perform pose estimation. A recent study emphasizes the need for precise estimation of pose keypoint coordinates and their significance in achieving accurate HAR \cite{9677850}. Therefore, we use ViTPose \cite{Xu2022ViTPoseSV}, as our method for pose estimation. Our choice of ViTPose is due to its lightweight decoder and Vision Transformer architecture, which have demonstrated state-of-the-art performance on the MS COCO dataset. We input to ViTPose the bounding box coordinates obtained with YOLOv5x corresponding to the detected human in a given frame $t$, and it returns a pose tensor $V^t$. For a sequence of frames, the ViTPose model generates a set of pose tensors as a skeleton sequence.
\begin{figure}[b]    \includegraphics[width=0.5\textwidth]{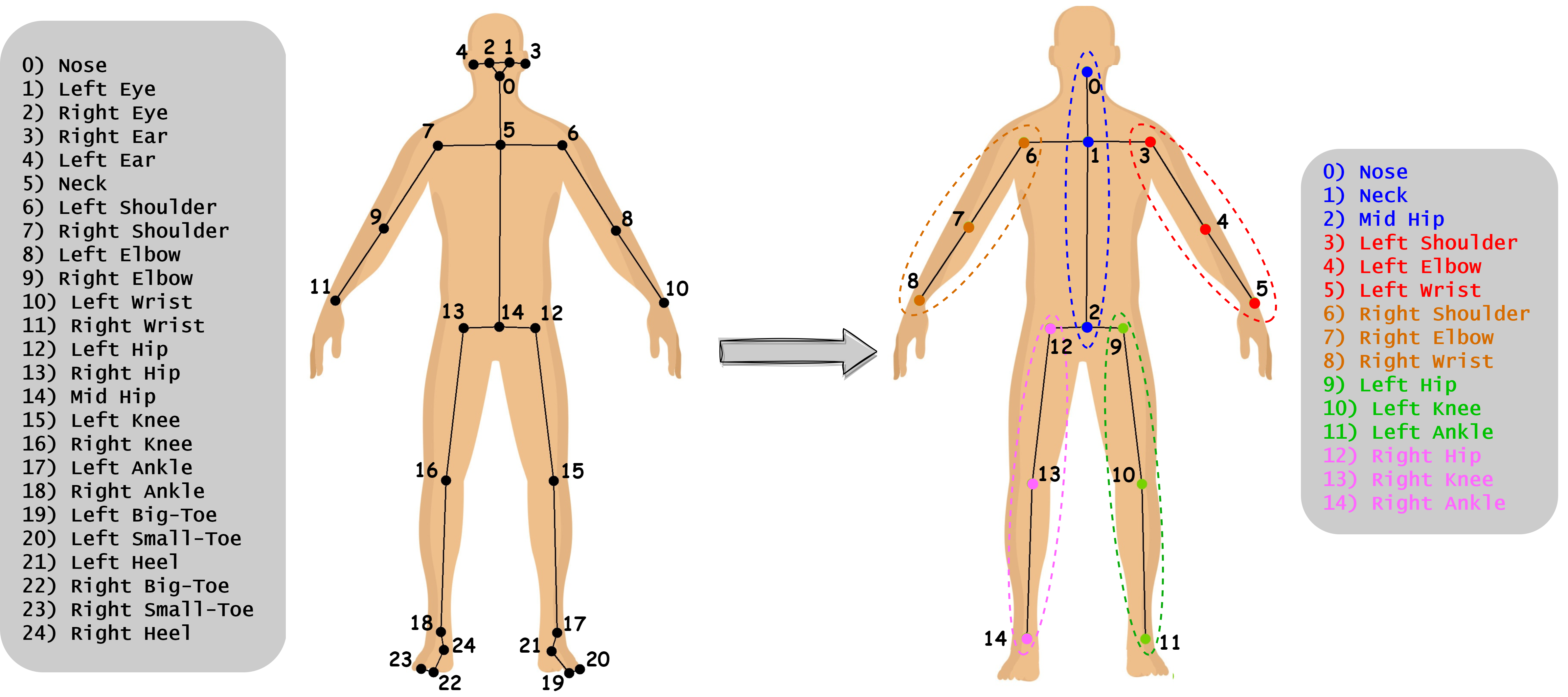}
\centering
	\caption{The human pose is a set of 2D connected keypoints corresponding to joints and important anatomic structures. (Left) Output of ViTPose model. (Right) Skeletons after our Selection-Permutation strategy.}
	\label{FIG:2}
\end{figure}

\begin{figure*}[t]
        \centering	\includegraphics[width=1.0\textwidth]{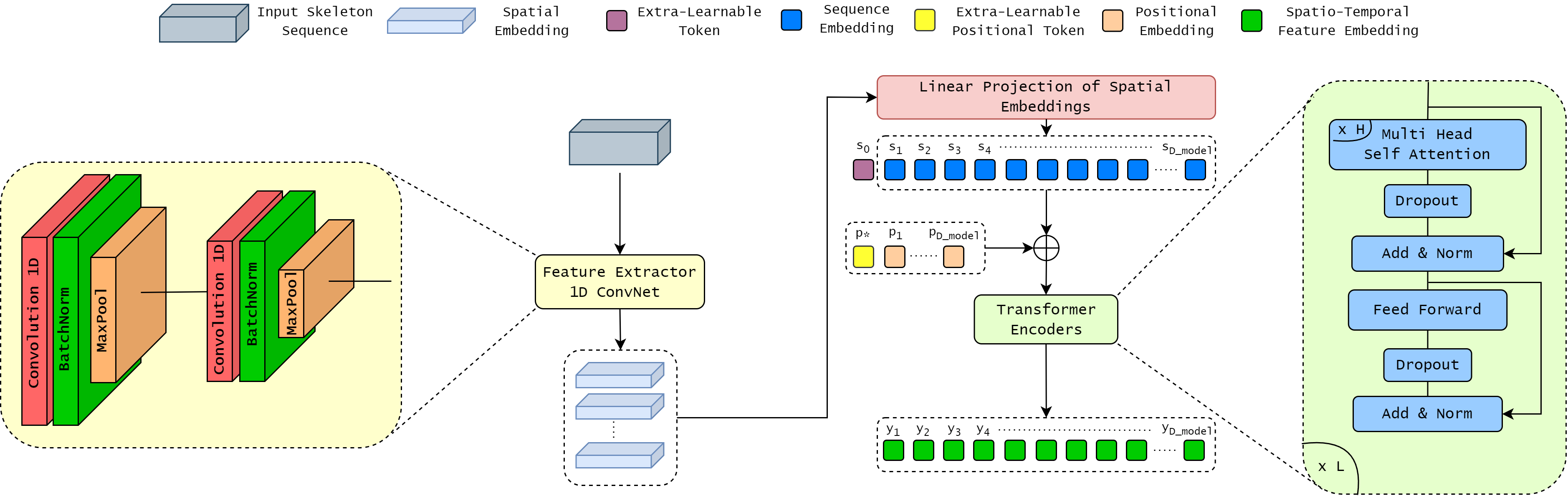}
	\caption{Illustration of our  Convolutional transformer model for spatio-temporal feature extraction. An input skeleton sequence is first fed to a 1D-ConvNet. The computed spatial embeddings are then fed to a transformer that generates the final spatio-temporal feature embeddings used for classification.}
	\label{FIG:3}
\end{figure*}

\subsection{Input skeleton pre-processing and Selection-Permutation strategy}
\label{3.2}
For skeleton-based HAR, previous studies have employed a methodology involving a limited number of long temporal skeleton sequences as input \cite{Yan2018SpatialTG,Shi2018TwoStreamAG}. These studies utilize a straightforward matching mechanism based on confidence scores to link skeletons into long sequences that cover the entire video. However, in ReL-SAR, we propose to sample a large number 
of short temporal skeleton sequences as input for better input consistency. ReL-SAR is therefore not dependent on the video length and does not necessitate zero-padding. By using multiple short video sequences, we can handle inputs with different lengths.
The ViTPose model outputs $25$ joint locations in the form of $(x,y)$ coordinates, along with a confidence score, resulting in a $25 \times 3$ tensor. The pose tensors generated by ViTPose capture the spatial information of the body pose keypoints across an entire video sequence of length $T$. We employed a set $X$ containing pose tensors $V^1, V^2, V^3, \ldots, V^T$ as input. Each $V^t$, $t\in \{1,.., T\}$, is a 2-dimensional tensor with only the spatial coordinates of the body pose keypoints in frame $t$.  Since humans can have various sizes in a frame, some joints may not always be extracted reliably. Therefore, we propose to select only $15$ essential joints that are easier to detect accurately. These $15$ joints were empirically selected among the original $25$ provided by the ViTPose model. Specifically, we remove the joints corresponding to eyes, ears, nose, toes, and heels. Each body is thus represented by a $15 \times 2$ tensor.

To effectively capture spatial relationships, we leverage a strategy that entails ordering the joints by permuting them in such a way that they are grouped by body parts. This ordering takes into account the inherent structure of the human body by partitioning the human skeleton into five anatomical regions based on the physical topology of the body as shown in Figure \ref{FIG:2}. This partitioning, which we believe to be intuitive, aims to capitalize on the insight that connections between proximate body parts within the input sequence can offer more nuanced and informative descriptions for the action representation. In fact, by organizing joints in a way that reflects the body's anatomical structure, we facilitate the extraction of local information in the convolutional layers of our model. This approach contrasts with a comprehensive attention focus on all joints simultaneously. Our strategy ensures a meaningful arrangement of joints that boosts the model ability to understand and interpret the spatial details present in the skeletal data. 
\subsection{Spatio-temporal feature extraction}
\label{3.3}

The model employed in ReL-SAR for spatio-temporal feature extraction is illustrated in Figure \ref{FIG:3}. Firstly, an input skeleton sequence $X$ goes through a 1D ConvNet encoder consisting of two blocks, each containing a convolutional 1D layer with $F$ filters, $K$ kernels, and a SeLU activation followed by a Batch-Normalization layer and a MaxPooling layer to extract spatial embeddings. The resulting spatial embeddings are then mapped to a dimension $D_{model}$ using a linear projection to obtain a sequence embedding $s_i$ that will serve as input to the Transformer encoder. We add an extra learnable token to the sequence embeddings following Dosovitskiy \cite{dosovitskiy2021an}. The role of this token is to aggregate information from the entire sequence. We refer to this token as the class token $[CLS]$ for consistency with previous work \cite{Touvron2020TrainingDI}, even though it is not attached to any label nor supervision in our case as in \cite{Caron2021EmergingPI}. Moreover, positional information is incorporated into the sequence via a learnable positional embedding tensor $p_{\text {i}} \in \mathbb{R}^{ D_{\text {model }}}$ added to all token representations, providing positional information that enables the model to effectively process sequential data. 
Our Transformer encoder comprises ($L$) layers, each consisting of alternating multi-head self-attention and feed-forward blocks ($H$). After each block, dropout, layer normalization, and residual connections are systematically applied. Each feed-forward block is a multi-layer perceptron with two layers and GeLU non-linearity. The initial layer serves to increase the dimensionality from $D_{model}$ to $D_{mlp}=4 \cdot D_{model}$ while incorporating the activation function. Conversely, the second layer reduces the dimensionality, restoring it from $D_{mlp}$ to $D_{model}$. 
The multi-head $Q K V$ self-attention mechanism $(MSA)$ relies on a trainable associative memory using key-value vector pairs. Specifically, for the $l$-th layer of the Transformer encoder and the $h$-th attention head, the computation of queries ($Q$), keys ($K$), and values ($V$) is expressed as:
\begin{equation}
    Q=X W_Q\text { , }  K=X W_K \text { and } V=X W_V
\end{equation}
where $W_Q$, $W_K$ and $W_V$ belong to $\mathbb{R}^{D_{\text {model }} \times D_h}$, $D_h$ being the dimension of the attention head. For each self-attention head $(SA)$ and every element within the input sequence, a weighted sum is performed over all the corresponding values in ($V$). Finally, the outputs from all attention heads are concatenated, and they undergo a linear projection to revert to the original dimension $D_{model}$. This attention mechanism operates in the time domain, enabling the creation of a comprehensive spatio-temporal feature embedding representation by establishing connections across different time windows.

\begin{figure*}[t]
        \centering
	\includegraphics[width=0.8\textwidth]{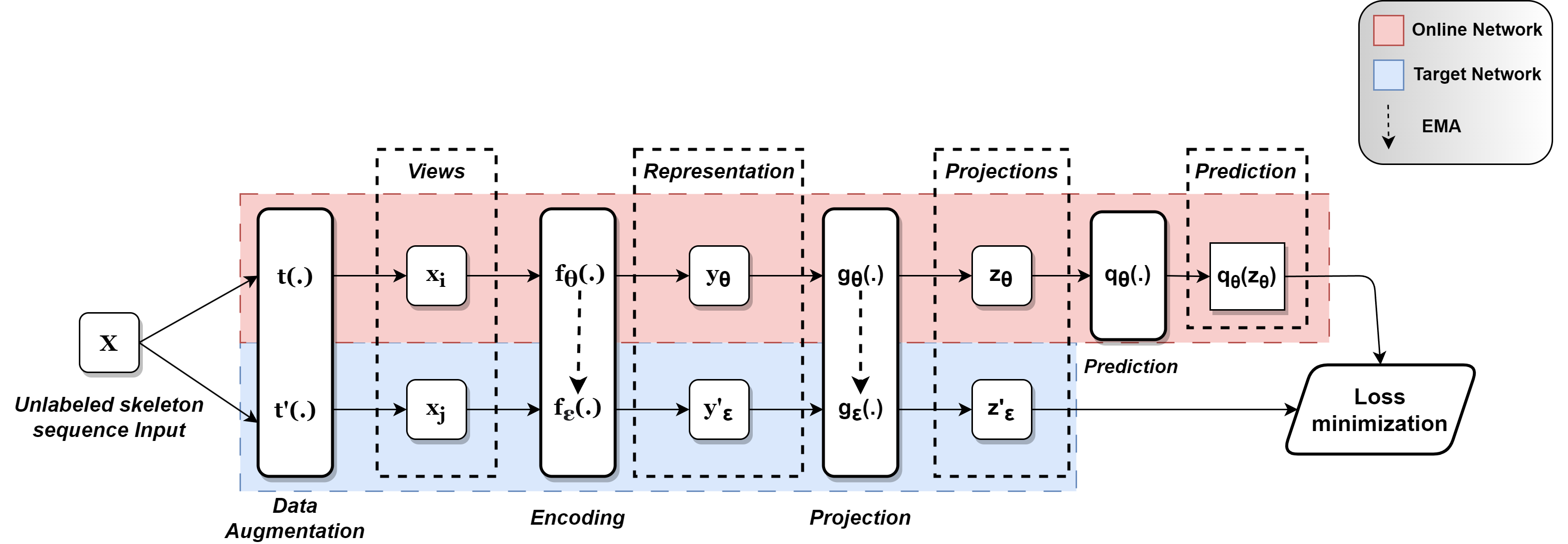}
	\caption{Proposed skeleton-based BYOL for action recognition.}
	\label{FIG:1}
\end{figure*}
\subsection{Bootstrap Your Own Latent (BYOL) for representation learning}\label{3.4}


In terms of modern self-supervision learning counterparts of BYOL, they use negative sampling and image (dis)similarity (SimCLR \cite{Chen2020ASF}, MoCo \cite{He2019MomentumCF}, DINO \cite{Caron2021EmergingPI}). They are strongly dependent on the tedious use of augmentation methods.
To apply BYOL, we use two neural networks having the same Convolutional Transformer architecture as presented in the previous section, referred to as online and target networks (see Figure \ref{FIG:1}). The online network is defined by a set of weights $\theta$, and the target network uses a different set of weights $\xi$. First, BYOL produces two augmented views, $x_i \triangleq t(x)$ and $x_j \triangleq t^{\prime}(x)$, from an unlabeled skeleton sequence input
$X$ by applying respectively data augmentations $t$ and
$t^{\prime}$. Then, the online network outputs
a representation $y_\theta$, a projection $z_\theta$, and a prediction $q_\theta\left(z_\theta\right)$
from the first view $x_i$. On the other hand, the target network
outputs $y_{\xi}^{\prime}$
and the target projection $z_{\xi}^{\prime}$
from the second view
$x_j$. Finally, the following mean squared error between the
L2-normalized predictions $\overline{q_\theta}\left(z_\theta\right)$ and target projections $\bar{z}_{\xi}^{\prime}$ is
calculated:
\begin{equation}
L_{\theta, \xi} \triangleq\left\|\overline{q_\theta}\left(z_\theta\right)-\bar{z}_{\xi}^{\prime}\right\|_2^2=2-2 \cdot \frac{\left\langle q_\theta\left(z_\theta\right), z_{\xi}^{\prime}\right\rangle}{\left\|q_\theta\left(z_\theta\right)\right\|_2 \cdot\left\|z_{\xi}^{\prime}\right\|_2},
\end{equation}
where $\langle\cdot, \cdot\rangle$ denotes the inner product. To make the loss symmetric, 
$L_{\theta, \xi}$, $L_{\theta, \xi}^{\prime}$ is computed by feeding $x_j$ to the online network and
$x_i$ to the target network. The final loss is defined as:
\begin{equation}
    L_{\theta, \xi}^{\mathrm{BYOL}}= L_{\theta, \xi}+L_{\theta, \xi}^{\prime},
\end{equation}
At each training step, BYOL minimizes this loss
function with respect to $\theta$ only, but $\xi$ is a slowly exponential moving average:
\begin{equation}
\theta: \xi \leftarrow \tau \xi+(1-\tau) \theta,
\end{equation}
where $\tau$ is a
target decay rate. Empirical evidence demonstrates that the combination of adding the predictor to the online network and representing the target network as an exponential moving average of the online network encourages richer representations while avoiding collapsed solutions. 


We incorporate light data augmentations in the process of training BYOL to learn representations that are semantically relevant for action recognition \cite{Grill2020BootstrapYO}. Transforming the skeleton input sequence creates variations that minimize overfitting and improve generalization for stronger performance. Our data augmentations used for BYOL are:

\begin{itemize}
    \item \textbf{Noise:} This transformation adds a random noise (or jitter) to joint coordinates, forcing the model to learn features invariant to minor skeleton variations.
    \item \textbf{Scaling:} This transformation changes the magnitude of the input sequence through multiplying with a random scaling factor, making the model more robust to variations in scale and offset of the skeleton sequence.
    \item \textbf{Vertical Flip:} This transformation applies a randomized vertical flipping to the input sequences, presenting the model with new temporal dynamics while maintaining feature continuity.

\end{itemize}

\subsection{Action Classification}
\label{3.5}
After BYOL training, we obtain two encoders that capture meaningful spatio-temporal features: $f_\theta$ derived from the online network and $f_{\xi}$ derived from the target network. As in Grill \emph{et al.}~\cite{Grill2020BootstrapYO}, only the encoder of the online network is kept as the final model. With the embeddings from the online encoder, we include a simple linear classifier $f(.)$ on top of the encoder to perform action recognition. Unlike other contrastive-based self-supervised skeleton action recognition methods \cite{Tang2020ExploringCL,Kim2022GloballocalMT}, which verify the model via linear evaluation protocol, we apply an end-to-end fine-tuning and focus on the evaluation of learned feature hierarchies for the skeleton action recognition tasks. 

\section{Experiments}
\label{experiments}

In this section, we conducted an extensive evaluation of ReL-SAR on several publicly available datasets for human activity recognition (HAR). We will describe the datasets and metrics used, and then present the implementation details followed by a detailed description and discussion of our experiments.

\subsection{Datasets and Evaluation Metric}

In our experiments, we evaluated the proposed approach on the following datasets, with a particular focus on their limited size: 
\textbf{Multi-Camera Action Dataset (MCAD) \cite{Li2016MultiCameraAD}:} It consists of 14,298 action samples, covering 18 action categories;
\textbf{INRIA Xmas Motion Acquisition Sequences dataset (IXMAS) \cite{Weinland2006FreeVA}:} It contains 12 action categories with 1800 action samples;
\textbf{Joint-annotated Human Motion DataBase (JHMDB) \cite{Jhuang2013TowardsUA}:} It consists of 960 video sequences depicting 21 actions;
\textbf{Northwestern-UCLA (NW-UCLA) \cite{Wang2014CrossViewAM}: }It contains 1484 action samples covering ten action classes. We adopt the same evaluation benchmark in \cite{Wang2014CrossViewAM}; cameras 1 and 2 capture samples as training data and the rest as testing data.
On all the datasets, classification accuracy is used as an evaluation metric. 

\subsection{Implementation and Training Details}

All experiments were performed using an Nvidia GeForce RTX 3060 GPU. We implemented ReL-SAR using TensorFlow. For human detection, YOLOv5x \cite{https://doi.org/10.5281/zenodo.3908559} with pre-trained weights from MS COCO is employed. For human pose estimation, we used the ViTPose model \cite{Xu2022ViTPoseSV}, specifically, the huge model trained on MS COCO. In our implementation, the specific value of $T$ is dataset-dependent. We selected it based on the length of the videos to ensure sufficient coverage of important action information. We chose $T$= 30 for MCAD, IXMAS, and NW-UCLA datasets, and $T$= 12 for JHMDB dataset. We have set the parameters of our Convolutional Transformer as follows: $F$= 192, $K$= 3, $H$= 3, $L$= 6, and $D_{\text {model}}$= 192 with $F$ and $K$ being the number of filters and kernel size respectively of each Conv1D layer. $L$ is the number of Transformer layers, and $H$ is the number of heads.

\subsubsection{BYOL pre-training} We trained BYOL with unlabeled skeleton sequences for 100 epochs with a batch size of 64. We used stochastic gradient descent (SGD) with a weight decay of 1e-4 and a momentum of 0.9. We opted for a cosine decay learning rate schedule with an initial rate of 1e-2 and 1000 decay steps.

\subsubsection{Fully Supervised baseline training} 
The fully supervised baseline corresponds to our convolutional transformer architecture without fine-tuning or freezing BYOL encoder.
We trained this model for 500 epochs with a batch size of 128. The AdamW optimization algorithm is employed with a step drop of the learning rate $\lambda$ to 1e-4 at a fixed percentage (80\%) of the total number of epochs. A linear warmup was performed for the first 200 epochs, representing 40\% of training. Weight decay of 1e-4 and label smoothing of 0.1 were used for regularization.

\subsection{Evaluating the Effectiveness of Learned Feature Hierarchies}

To assess the quality of the learned representations, we conducted an analysis evaluating the performance of various fine-tuning scenarios in Table \ref{tab:tab1}. In the fully fine-tuning setting, we first trained BYOL with unlabeled skeleton sequences and then fine-tuned with the linear classifier in a supervised manner. In the frozen settings, we preserve the weights of the specified early convolutional layer learned with BYOL, and only fine-tune the weights of the later, more task-related layers.

This analysis aimed to determine whether the features extracted from various layers trained in a self-supervised manner with BYOL exhibited differences in quality compared to their fully supervised counterparts with respect to the end-task performance, and if so, which layer should be utilized for this purpose.  Table \ref{tab:tab1} provides action recognition accuracy results. Fine-tuning the entire model after applying BYOL yielded improvements over the supervised baseline across all four datasets. This highlights the ability of BYOL to achieve more representative feature extraction. Freezing the first Conv1D layer obtained from BYOL and fine-tuning the remainder yielded optimal performance on the MCAD, JHMDB, and NW-UCLA datasets. We achieved a 2.09\% action recognition accuracy boost over the supervised baseline on NW-UCLA dataset, from 93.18\% to
95.27\%, when freezing Conv1D Layer1 of BYOL. This suggests that the low-level information captured by BYOL generalizes more effectively compared to the supervised baseline. Interestingly, freezing Conv1D layer 2 resulted in lower performance compared to freezing Conv1D layer 1 across all datasets. This could be attributed to the second layer representations becoming overly specific to data augmentations. However, freezing Conv1D layer 2 still demonstrated improvements in performance and generalization compared to the supervised baseline. These findings underline the ability of BYOL to learn more general skeleton features in the absence of explicit labels.

\begin{table}[t]
    \scriptsize
    \centering
    \caption{Evaluating the effectiveness of learned features. Performance comparison with the fully-supervised baseline on all four datasets. We present the action recognition accuracy (\%). Best result is in bold and the second best result is in italics.}
\begin{tabular}{lccccc}
\hline & \multicolumn{4}{c}{ \textbf{Dataset} } \\
\textbf{Model} & MCAD & JHMDB & IXMAS & NW-UCLA \\
\hline Fully Supervised baseline & $87.64$ & $75.57$ & $88.65$ & $93.18$ \\
\hline Fully Fine-tuning & $\textit{89.47}$ & $\textit{80.36}$ & $\mathbf{91.78}$ & $\textit{93.92}$ \\
Freezing Conv1D Layer1 & $\mathbf{90.96}$ & $\mathbf{80.48}$ & $\textit{90.75}$ & $\mathbf{95.27}$  \\
Freezing Conv1D Layer2 & $88.95$ & $78.44$ & $90.87$ & $91.89$  \\
\hline
\end{tabular}
\label{tab:tab1}
\end{table}

\begin{table*}[t]
    \scriptsize
    \centering
\caption{Comparisons of action recognition results against semi-supervised learning approaches on NW-UCLA dataset. V./ C. denotes the number of labeled videos per class. Best results are in bold and the second best results are in italics.}
\begin{tabular}{ccccccc}
\hline \textbf{Method} & \textbf{1 V./ C.} & \textbf{5 V./ C.} & \textbf{10 V./ C.} & \textbf{15 V./ C.} & \textbf{30 V./ C.} & \textbf{40 V./ C.} \\
\hline  Pseudolabels \cite{Lee2013PseudoLabelT}  & $-$ & $35.6$ & $-$ & $48.9$ & $60.6$ & $65.7$ \\
 VAT \cite{Miyato2017VirtualAT} & $-$ & $44.8$ & $-$ & $63.8$ & $73.7$ & $73.9$ \\
$\mathrm{S^4} \mathrm{L}$ (Inpainting) \cite{Zhai2019S4LSS} & $-$ & $35.3$ & $-$ & $46.6$ & $54.5$ & $60.6$ \\
 ASSL \cite{Si2020AdversarialSL} & $-$ & $52.6$ & $-$ & $74.8$ & $78.0$ & $78.4$ \\
 LongT GAN \cite{Zheng2018UnsupervisedRL} & $18.3$ & $-$ & $59.9$ & $-$ & $-$ & $-$ \\
 $\mathrm{MS^2} \mathrm{L}$ \cite{Lin2020MS2LMS} & $21.3$ & $-$ & $60.5$ & $-$ & $-$ & $-$ \\
'$\mathrm{TS+SS}$' Colorization \cite{Yang2021SkeletonCC}  & $\textit{41.9}$ & $\textit{57.2}$ & $\textbf{75.0}$ & $\textit{76.0}$ & $\textit{83.0}$ & $\textit{84.9}$ \\
\hline
ReL-SAR (Ours)  & \textbf{43.2} & \textbf{61.3} & $\textit{72.7}$ & $\textbf{76.5}$ & $\textbf{83.9}$ & $\textbf{87.1}$ \\
\hline
\end{tabular}
\label{tab:tab3}
\end{table*}

\subsection{Semi-Supervised Evaluation}
Under the semi-supervised setting, we initially pre-trained the online encoder $f_\theta$ of BYOL with unlabeled skeleton sequences. We then fine-tune the linear classifier $f(.)$ with a small ratio of action annotations in a supervised fashion. 
We compare ReL-SAR with other state-of-the-art methods in the same semi-supervised setting for NW-UCLA dataset. Results are given in Table \ref{tab:tab3}. Following Yang \emph{et al.}~\cite{Yang2021SkeletonCC}, we derived labeled data by uniformly sampling 1, 5, 10, 15, 30, and 40 videos from the training set. As Table \ref{tab:tab3} shows, ReL-SAR performs better than the state-of-the-art on the NW-UCLA dataset when less training data is available.

\subsection{Impact of the proposed Selection-Permutation strategy}


\begin{figure}[t]
        \centering
\includegraphics[width=0.48\textwidth]{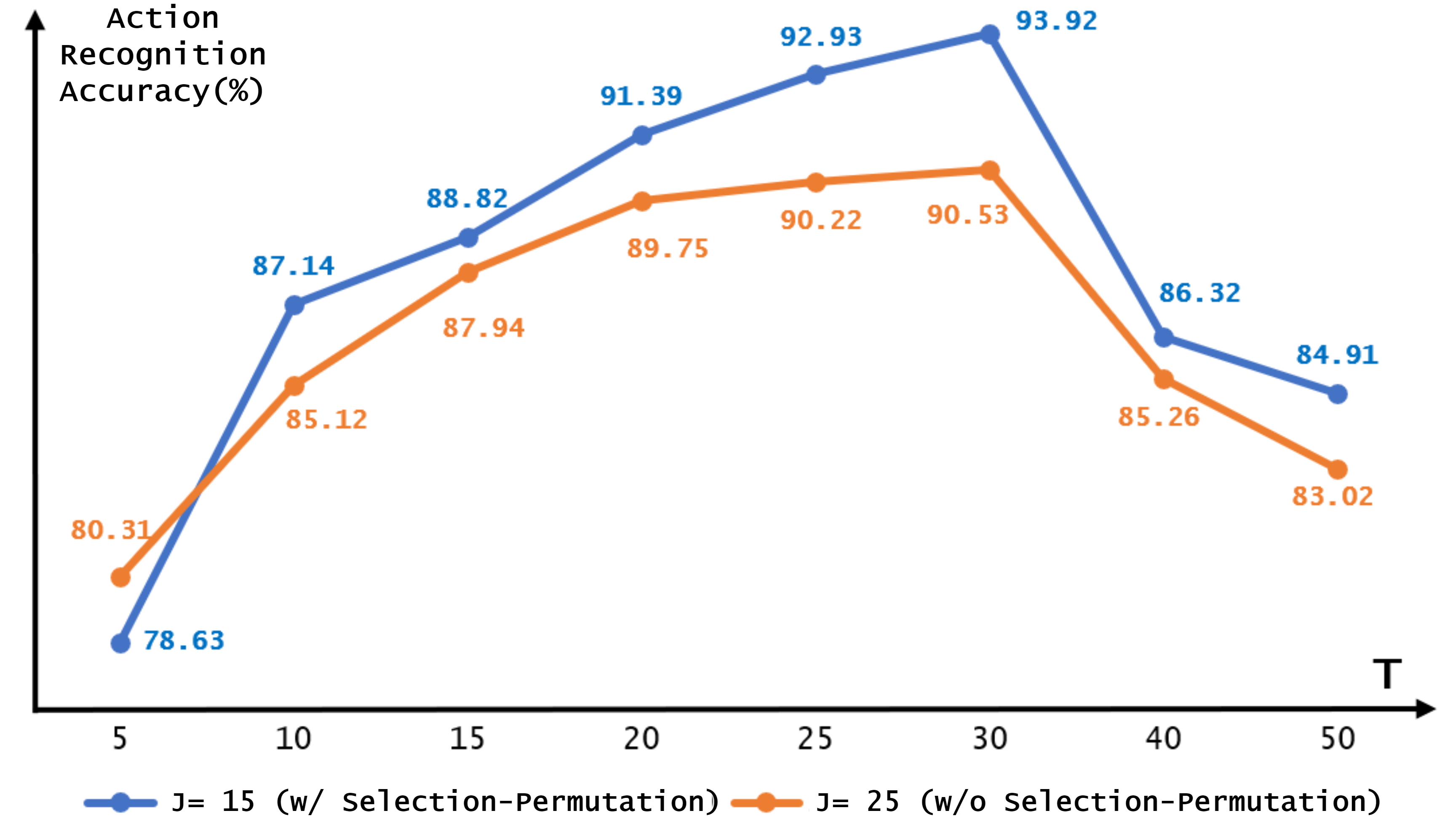}
	\caption{We fix the length $T$ of the input sequence and test the influence of our Selection-Permutation strategy. Action recognition accuracy of fully fine-tuning is visualized.}
	\label{FIG:5}
\end{figure}

We showcase the advantages of our input skeleton structure, highlighting improvements in HAR. 
In ReL-SAR, we used a large number of short skeleton sequences of length $T$ as input for skeleton action recognition. To investigate the effect of sequence length $T$, we tested multiple combinations of sequence length and number of joints. Results of the different combinations are given in Figure \ref{FIG:5}.
Interestingly, we find that the highest performances are obtained when using our Selection-Permutation strategy with $J$= $15$ except for $T$= $5$. Using this strategy, we achieve the best result (93.92\% accuracy) with $T$= $30$. We also observe that with the increasing size of sequence, the performance of each model increases until a certain length, after which the recognition accuracy slightly decreases. This suggests an optimal balance between sequence length and joint number for achieving peak performance in skeleton action recognition.
\subsection{Comparison to the State-of-the-Art}
To demonstrate the benefits of ReL-SAR, we conducted comparisons with state-of-the-art methods on four datasets: IXMAS, MCAD, JHMDB, and NW-UCLA. The results are summarized in Tables \ref{tab:tab6}, \ref{tab:tab7}, \ref{tab:tab8} and \ref{tab:tab9}, respectively. Top-1 Accuracy denotes the action recognition accuracy, Param denotes the number of parameters, and FLOPs denotes the Floating point Operations per Second. Best result is bolded. Second Best result is in italics.
ReL-SAR is ranked first in action recognition accuracy on MCAD, IXMAS, and JHMDB, outperforming other state-of-the-art methods. On NW-UCLA datasets, although our method does not exceed the highest action recognition accuracy, it achieves comparable performance while being much less computationally expensive than the best performing method. Specifically, ReL-SAR obtains comparable accuracy of 95.27\% with fewer parameters (2.80M vs 10.08M) and 53x fewer FLOPs (0.18G vs 9.60G) compared to the top performing HD-GCN \cite{Lee2022HierarchicallyDG}. This substantial reduction in computational requirements enables more efficient and practical deployment for real-time action recognition applications. In fact, Graph Convolutional Networks (GCN) have emerged as a popular approach for achieving high performance in skeleton-based action recognition. These GCN-based approaches often require a large computational cost. ReL-SAR is based on a reasonable design of a lightweight convolutional transformer and achieves better or similar recognition accuracy as the GCN-based methods with less computational cost by relying on representations learned on a self supervised manner with BYOL, which is advantageous for practical implementation solutions.

\begin{table}[]
    \scriptsize
   \centering
  \caption{Comparison of ReL-SAR against state-of-the-art on the IXMAS dataset. Top-1 Accuracy denotes the action recognition accuracy and Param denotes the number of parameters. Best result is bolded. Second Best result is in italic.}
\begin{tabular}{l|c|c}
\hline \textbf{Methods} & \textbf{Top-1 Accuracy (\%)} & \textbf{Param. (M)} \\
\hline MBP/VLBP \cite{Baumann2016RecognizingHA} & $80.55$ & --\\
 HOMID \cite{Chun2016HumanAR} & $83.03$ & --\\
T-VLAD \cite{Naeem2021TVLADTV} & $84.80$ & -- \\
SDEG \cite{Vishwakarma2016APU} & $85.80$ & --\\
H-VLBP + DSAutEnc \cite{Kiruba2019HexagonalVL} & $88.76$ & -- \\
CCA \cite{Kiran2021MultiLayeredDL} & $89.60$ & --\\
Rehman et.al \cite{Malik2023MultiViewHA} & $\textit{90.97}$ & --\\
\hline
\hline ReL-SAR (Ours) & $\mathbf{91.78}$ & $\mathbf{2.80}$\\

\hline
\end{tabular}
 \label{tab:tab6}%
\end{table}
\begin{table}[]
    \scriptsize
   \centering
  \caption{Comparison of ReL-SAR against state-of-the-art on the MCAD dataset. Best result is bolded. Second Best result is in italic.}
\begin{tabular}{l|c|c}
\hline \textbf{Methods} & \textbf{Top-1 Accuracy (\%)} & \textbf{Param. (M)} \\
\hline
Covariance Matrices \cite{iet:/content/journals/10.1049/iet-cvi.2014.0018} & $64.30$ & -- \\
T-VLAD \cite{Naeem2021TVLADTV} & $78.60$ & -- \\
Wenhui et al. \cite{Li2016MultiCameraAD} & $81.70$ & --\\
Conflux LSTM \cite{Ullah2020ConfluxLN} & $86.90$ & $6.80$\\
Rehman et al. \cite{Malik2023MultiViewHA} & $\textit{89.70}$ & --\\
\hline
\hline ReL-SAR (Ours) & $\mathbf{90.96}$ & $\mathbf{2.80}$\\
\hline
\end{tabular}
 \label{tab:tab7}%
\end{table}
\begin{table}[t]
    \scriptsize
   \centering
  \caption{Comparison of ReL-SAR against state-of-the-art on the JHMDB dataset. Best result is bolded. Second Best result is in italic.}
\begin{tabular}{l|c|c}
\hline \textbf{Methods} & \textbf{Top-1 Accuracy (\%)} & \textbf{Param. (M)}\\
\hline
FCN \cite{Zolfaghari2017ChainedMN} & $56.80$ & $17.50$ \\
PoTion \cite{Choutas2018PoTionPM} & $59.10$ & $4.87$ \\
Two Branches LSTM \cite{Li2022TowardsBA} & $61.24$ & --\\
EHPI \cite{Yang2019MakeSA} & $65.50$ & $\mathbf{1.22}$\\
JMRN \cite{Shah2020PoseAJ} & $68.55$ & --\\
DD-Net \cite{Yang2019MakeSA}  & $77.20$ & $1.82$\\
TD-Net \cite{Nguyen2022ARA}  & $79.30$ & $1.88$\\
BiLSTM + DCNN \cite{Muhammad2021HumanAR} & $\textit{80.20}$ & --\\
\hline
\hline ReL-SAR (Ours) & $\mathbf{80.48}$ & $2.80$\\
\hline
\end{tabular}
 \label{tab:tab8}%
\end{table}
\begin{table}[t]
    \scriptsize
   \centering
  \caption{Comparison of ReL-SAR against state-of-the-art on the NW-UCLA dataset. FLOPs denotes the Floating point Operations per Second. Best result is bolded. Second Best result is in italic.}
\begin{tabular}{l|c|c|c}
\hline \textbf{Methods} & \textbf{Top-1 Acc (\%)} & \textbf{Param. (M)} &\textbf{FLOPs (G)} \\
\hline
1 s MSSTNet \cite{Cheng2023MultiscaleSC} & $92.20$ & $-$ & $0.20$\\
SGN \cite{Zhang2019SemanticsGuidedNN} & $92.50$ & $-$ & $-$\\
2 s AGC-LSTM  \cite{Si2019AnAE}  & $93.30$ & $45.70$ & $10.90$\\
Shift-GCN \cite{Cheng2020SkeletonBasedAR} & $94.60$ & $2.80$ & $0.70$\\
DC-GCN+ADG \cite{Cheng2020DecouplingGW} & $95.30$ & $3.08$ & $64.80$\\
InfoGCN \cite{Chi2022InfoGCNRL} & $\textit{97.00}$ & $9.42$ & $10.08$\\
HD-GCN \cite{Lee2022HierarchicallyDG} & $\mathbf{97.20}$ & $10.08$ & $9.60$\\
\hline
\hline ReL-SAR (Ours) & $95.27$ & $\mathbf{2.80}$ & $\mathbf{0.18}$\\
\hline
\end{tabular}
 \label{tab:tab9}%
\end{table}

\section{Conclusion}
We introduced ReL-SAR, a lightweight convolutional transformer model for skeleton-based action recognition. It relies on a Selection-Permutation strategy to obtain informative joint-level inputs and a self-supervised pre-training with BYOL for learning robust low-level skeleton features without labels. The experiments on four datasets demonstrate that ReL-SAR achieves superior action recognition performance while being computationally efficient. Overall, ReL-SAR provides an accurate yet lightweight approach for skeleton-based action recognition, making it suitable for deployment on limited-resource devices.


\bibliographystyle{IEEEtran}
\bibliography{conference}

\begin{thebibliography}{10}
\providecommand{\url}[1]{#1}
\csname url@samestyle\endcsname
\providecommand{\newblock}{\relax}
\providecommand{\bibinfo}[2]{#2}
\providecommand{\BIBentrySTDinterwordspacing}{\spaceskip=0pt\relax}
\providecommand{\BIBentryALTinterwordstretchfactor}{4}
\providecommand{\BIBentryALTinterwordspacing}{\spaceskip=\fontdimen2\font plus
\BIBentryALTinterwordstretchfactor\fontdimen3\font minus \fontdimen4\font\relax}
\providecommand{\BIBforeignlanguage}[2]{{%
\expandafter\ifx\csname l@#1\endcsname\relax
\typeout{** WARNING: IEEEtran.bst: No hyphenation pattern has been}%
\typeout{** loaded for the language `#1'. Using the pattern for}%
\typeout{** the default language instead.}%
\else
\language=\csname l@#1\endcsname
\fi
#2}}
\providecommand{\BIBdecl}{\relax}
\BIBdecl

\bibitem{Zheng2018UnsupervisedRL}
N.~Zheng, J.~Wen, R.~Liu, L.~Long, J.~Dai, and Z.~Gong, ``Unsupervised representation learning with long-term dynamics for skeleton based action recognition,'' in \emph{AAAI Conference on Artificial Intelligence}, 2018.

\bibitem{Su2019PREDICTC}
K.~Su, X.~Liu, and E.~Shlizerman, ``Predict \& cluster: Unsupervised skeleton based action recognition,'' \emph{2020 IEEE/CVF Conference on Computer Vision and Pattern Recognition (CVPR)}, pp. 9628--9637, 2019.

\bibitem{Lin2020MS2LMS}
L.~Lin, S.~Song, W.~Yang, and J.~Liu, ``Ms2l: Multi-task self-supervised learning for skeleton based action recognition,'' \emph{Proceedings of the 28th ACM International Conference on Multimedia}, 2020.

\bibitem{Rao2020AugmentedSB}
H.~Rao, S.~Xu, X.~Hu, J.~Cheng, and B.~Hu, ``Augmented skeleton based contrastive action learning with momentum lstm for unsupervised action recognition,'' \emph{Inf. Sci.}, vol. 569, pp. 90--109, 2020.

\bibitem{Grill2020BootstrapYO}
J.-B. Grill, F.~Strub, F.~Altch'e, C.~Tallec, P.~H. Richemond, E.~Buchatskaya, C.~Doersch, B.~{\'A}. Pires, Z.~D. Guo, M.~G. Azar, B.~Piot, K.~Kavukcuoglu, R.~Munos, and M.~Valko, ``Bootstrap your own latent: A new approach to self-supervised learning,'' \emph{ArXiv}, vol. abs/2006.07733, 2020.

\bibitem{7299176}
B.~Fernando, E.~Gavves, M.~José~Oramas, A.~Ghodrati, and T.~Tuytelaars, ``Modeling video evolution for action recognition,'' in \emph{2015 IEEE Conference on Computer Vision and Pattern Recognition (CVPR)}, 2015, pp. 5378--5387.

\bibitem{7298714}
Y.~Du, W.~Wang, and L.~Wang, ``Hierarchical recurrent neural network for skeleton based action recognition,'' in \emph{2015 IEEE Conference on Computer Vision and Pattern Recognition (CVPR)}, 2015, pp. 1110--1118.

\bibitem{Liu2017EnhancedSV}
M.~Liu, H.~Liu, and C.~Chen, ``Enhanced skeleton visualization for view invariant human action recognition,'' \emph{Pattern Recognit.}, vol.~68, pp. 346--362, 2017.

\bibitem{Yang2019MakeSA}
F.~Yang, S.~Sakti, Y.~Wu, and S.~Nakamura, ``Make skeleton-based action recognition model smaller, faster and better,'' \emph{Proceedings of the ACM Multimedia Asia}, 2019.

\bibitem{Choutas2018PoTionPM}
V.~Choutas, P.~Weinzaepfel, J.~Revaud, and C.~Schmid, ``Potion: Pose motion representation for action recognition,'' \emph{2018 IEEE/CVF Conference on Computer Vision and Pattern Recognition}, pp. 7024--7033, 2018.

\bibitem{dosovitskiy2021an}
\BIBentryALTinterwordspacing
A.~Dosovitskiy, L.~Beyer, A.~Kolesnikov, D.~Weissenborn, X.~Zhai, T.~Unterthiner, M.~Dehghani, M.~Minderer, G.~Heigold, S.~Gelly, J.~Uszkoreit, and N.~Houlsby, ``An image is worth 16x16 words: Transformers for image recognition at scale,'' in \emph{International Conference on Learning Representations}, 2021. [Online]. Available: \url{https://openreview.net/forum?id=YicbFdNTTy}
\BIBentrySTDinterwordspacing

\bibitem{Kundu2018UnsupervisedFL}
J.~N. Kundu, M.~Gor, P.~K. Uppala, and R.~V. Babu, ``Unsupervised feature learning of human actions as trajectories in pose embedding manifold,'' \emph{2019 IEEE Winter Conference on Applications of Computer Vision (WACV)}, pp. 1459--1467, 2018.

\bibitem{Thoker2021SkeletonContrastive3A}
F.~M. Thoker, H.~Doughty, and C.~G.~M. Snoek, ``Skeleton-contrastive 3d action representation learning,'' \emph{Proceedings of the 29th ACM International Conference on Multimedia}, 2021.

\bibitem{Li20213DHA}
L.~Li, M.~Wang, B.~Ni, H.~Wang, J.~Yang, and W.~Zhang, ``3d human action representation learning via cross-view consistency pursuit,'' \emph{2021 IEEE/CVF Conference on Computer Vision and Pattern Recognition (CVPR)}, pp. 4739--4748, 2021.

\bibitem{https://doi.org/10.5281/zenodo.3908559}
\BIBentryALTinterwordspacing
G.~Jocher, ``ultralytics/yolov5,'' 2020. [Online]. Available: \url{https://github.com/ultralytics/yolov5}
\BIBentrySTDinterwordspacing

\bibitem{9677850}
S.~Mroz, N.~Baddour, C.~McGuirk, P.~Juneau, A.~Tu, K.~Cheung, and E.~Lemaire, ``Comparing the quality of human pose estimation with blazepose or openpose,'' in \emph{2021 4th International Conference on Bio-Engineering for Smart Technologies (BioSMART)}, 2021, pp. 1--4.

\bibitem{Xu2022ViTPoseSV}
Y.~Xu, J.~Zhang, Q.~Zhang, and D.~Tao, ``Vitpose: Simple vision transformer baselines for human pose estimation,'' \emph{Advances in Neural Information Processing Systems, 2022}, 2022.

\bibitem{Yan2018SpatialTG}
S.~Yan, Y.~Xiong, and D.~Lin, ``Spatial temporal graph convolutional networks for skeleton-based action recognition,'' in \emph{AAAI Conference on Artificial Intelligence}, 2018.

\bibitem{Shi2018TwoStreamAG}
L.~Shi, Y.~Zhang, J.~Cheng, and H.~Lu, ``Two-stream adaptive graph convolutional networks for skeleton-based action recognition,'' \emph{2019 IEEE/CVF Conference on Computer Vision and Pattern Recognition (CVPR)}, pp. 12\,018--12\,027, 2018.

\bibitem{Touvron2020TrainingDI}
H.~Touvron, M.~Cord, M.~Douze, F.~Massa, A.~Sablayrolles, and H.~J'egou, ``Training data-efficient image transformers \& distillation through attention,'' in \emph{International Conference on Machine Learning}, 2020.

\bibitem{Caron2021EmergingPI}
M.~Caron, H.~Touvron, I.~Misra, H.~J'egou, J.~Mairal, P.~Bojanowski, and A.~Joulin, ``Emerging properties in self-supervised vision transformers,'' \emph{2021 IEEE/CVF International Conference on Computer Vision (ICCV)}, pp. 9630--9640, 2021.

\bibitem{Chen2020ASF}
T.~Chen, S.~Kornblith, M.~Norouzi, and G.~E. Hinton, ``A simple framework for contrastive learning of visual representations,'' \emph{ArXiv}, vol. abs/2002.05709, 2020.

\bibitem{He2019MomentumCF}
K.~He, H.~Fan, Y.~Wu, S.~Xie, and R.~B. Girshick, ``Momentum contrast for unsupervised visual representation learning,'' \emph{2020 IEEE/CVF Conference on Computer Vision and Pattern Recognition (CVPR)}, pp. 9726--9735, 2019.

\bibitem{Tang2020ExploringCL}
C.~I. Tang, I.~Perez-Pozuelo, D.~Spathis, and C.~Mascolo, ``Exploring contrastive learning in human activity recognition for healthcare,'' \emph{ArXiv}, vol. abs/2011.11542, 2020.

\bibitem{Kim2022GloballocalMT}
B.~Kim, H.~J. Chang, J.~Kim, and J.~Y. Choi, ``Global-local motion transformer for unsupervised skeleton-based action learning,'' in \emph{European Conference on Computer Vision}, 2022.

\bibitem{Li2016MultiCameraAD}
W.~Li, Y.~Wong, A.~Liu, Y.~Li, Y.~Su, and M.~Kankanhalli, ``Multi-camera action dataset for cross-camera action recognition benchmarking,'' \emph{2017 IEEE Winter Conference on Applications of Computer Vision (WACV)}, pp. 187--196, 2016.

\bibitem{Weinland2006FreeVA}
D.~Weinland, R.~Ronfard, and E.~Boyer, ``Free viewpoint action recognition using motion history volumes,'' \emph{Comput. Vis. Image Underst.}, vol. 104, pp. 249--257, 2006.

\bibitem{Jhuang2013TowardsUA}
H.~Jhuang, J.~Gall, S.~Zuffi, C.~Schmid, and M.~J. Black, ``Towards understanding action recognition,'' \emph{2013 IEEE International Conference on Computer Vision}, pp. 3192--3199, 2013.

\bibitem{Wang2014CrossViewAM}
J.~Wang, X.~Nie, Y.~Xia, Y.~Wu, and S.-C. Zhu, ``Cross-view action modeling, learning, and recognition,'' \emph{2014 IEEE Conference on Computer Vision and Pattern Recognition}, pp. 2649--2656, 2014.

\bibitem{Lee2013PseudoLabelT}
D.-H. Lee, ``Pseudo-label : The simple and efficient semi-supervised learning method for deep neural networks,'' 2013.

\bibitem{Miyato2017VirtualAT}
T.~Miyato, S.~ichi Maeda, M.~Koyama, and S.~Ishii, ``Virtual adversarial training: A regularization method for supervised and semi-supervised learning,'' \emph{IEEE Transactions on Pattern Analysis and Machine Intelligence}, vol.~41, pp. 1979--1993, 2017.

\bibitem{Zhai2019S4LSS}
X.~Zhai, A.~Oliver, A.~Kolesnikov, and L.~Beyer, ``S4l: Self-supervised semi-supervised learning,'' \emph{2019 IEEE/CVF International Conference on Computer Vision (ICCV)}, pp. 1476--1485, 2019.

\bibitem{Si2020AdversarialSL}
C.~Si, X.~Nie, W.~Wang, L.~Wang, T.~Tan, and J.~Feng, ``Adversarial self-supervised learning for semi-supervised 3d action recognition,'' \emph{ArXiv}, vol. abs/2007.05934, 2020.

\bibitem{Yang2021SkeletonCC}
S.~Yang, J.~Liu, S.~Lu, M.~H. Er, and A.~C. Kot, ``Skeleton cloud colorization for unsupervised 3d action representation learning,'' \emph{2021 IEEE/CVF International Conference on Computer Vision (ICCV)}, pp. 13\,403--13\,413, 2021.

\bibitem{Lee2022HierarchicallyDG}
J.~Lee, M.~Lee, D.~Lee, and S.~Lee, ``Hierarchically decomposed graph convolutional networks for skeleton-based action recognition,'' \emph{2023 IEEE/CVF International Conference on Computer Vision (ICCV)}, pp. 10\,410--10\,419, 2022.

\bibitem{Baumann2016RecognizingHA}
F.~Baumann, A.~Ehlers, B.~Rosenhahn, and J.~Liao, ``Recognizing human actions using novel space-time volume binary patterns,'' \emph{Neurocomputing}, vol. 173, pp. 54--63, 2016.

\bibitem{Chun2016HumanAR}
S.~Y. Chun and C.-S. Lee, ``Human action recognition using histogram of motion intensity and direction from multiple views,'' \emph{IET Comput. Vis.}, vol.~10, pp. 250--256, 2016.

\bibitem{Naeem2021TVLADTV}
H.~B. Naeem, F.~Murtaza, M.~H. Yousaf, and S.~A. Velast{\'i}n, ``T-vlad: Temporal vector of locally aggregated descriptor for multiview human action recognition,'' \emph{Pattern Recognit. Lett.}, vol. 148, pp. 22--28, 2021.

\bibitem{Vishwakarma2016APU}
D.~K. Vishwakarma, R.~Kapoor, and A.~Dhiman, ``A proposed unified framework for the recognition of human activity by exploiting the characteristics of action dynamics,'' \emph{Robotics Auton. Syst.}, vol.~77, pp. 25--38, 2016.

\bibitem{Kiruba2019HexagonalVL}
K.~Kiruba, D.~S. Elizabeth, and C.~S.~R. Raj, ``Hexagonal volume local binary pattern (h-vlbp) with deep stacked autoencoder for human action recognition,'' \emph{Cognitive Systems Research}, vol.~58, pp. 71--93, 2019.

\bibitem{Kiran2021MultiLayeredDL}
S.~K. et~al., ``Multi-layered deep learning features fusion for human action recognition,'' \emph{Computers, Materials \& Continua}, 2021.

\bibitem{Malik2023MultiViewHA}
N.~ur~Rehman~Malik, U.~U. Sheikh, S.~A.~R. Abu-Bakar, and A.~Channa, ``Multi-view human action recognition using skeleton based-fineknn with extraneous frame scrapping technique,'' \emph{Sensors (Basel, Switzerland)}, vol.~23, 2023.

\bibitem{iet:/content/journals/10.1049/iet-cvi.2014.0018}
M.~Faraki, ``\BIBforeignlanguage{English}{Log-euclidean bag of words for human action recognition},'' \emph{\BIBforeignlanguage{English}{IET Computer Vision}}, vol.~9, pp. 331--339(8), June 2015.

\bibitem{Ullah2020ConfluxLN}
A.~Ullah, K.~Muhammad, T.~Hussain, and S.~W. Baik, ``Conflux lstms network: A novel approach for multi-view action recognition,'' \emph{Neurocomputing}, vol. 435, pp. 321--329, 2020.

\bibitem{Zolfaghari2017ChainedMN}
M.~Zolfaghari, G.~L. Oliveira, N.~Sedaghat, and T.~Brox, ``Chained multi-stream networks exploiting pose, motion, and appearance for action classification and detection,'' \emph{2017 IEEE International Conference on Computer Vision (ICCV)}, pp. 2923--2932, 2017.

\bibitem{Li2022TowardsBA}
X.~Li, S.~Onie, M.~Liang, M.~S. Larsen, and A.~Sowmya, ``Towards building a visual behaviour analysis pipeline for suicide detection and prevention,'' \emph{Sensors (Basel, Switzerland)}, vol.~22, 2022.

\bibitem{Shah2020PoseAJ}
A.~B. Shah, S.~K. Mishra, A.~Bansal, J.-C. Chen, R.~Chellappa, and A.~Shrivastava, ``Pose and joint-aware action recognition,'' \emph{2022 IEEE/CVF Winter Conference on Applications of Computer Vision (WACV)}, pp. 141--151, 2020.

\bibitem{Nguyen2022ARA}
T.~Nguyen, D.-T. Pham, H.~Vu, and T.-L. Le, ``A robust and efficient method for skeleton-based human action recognition and its application for cross-dataset evaluation,'' \emph{IET Comput. Vis.}, vol.~16, pp. 709--726, 2022.

\bibitem{Muhammad2021HumanAR}
K.~Muhammad, Mustaqeem, A.~Ullah, A.~S. Imran, M.~Sajjad, M.~S. Kiran, G.~Sannino, and V.~H. Albuquerque, ``Human action recognition using attention based lstm network with dilated cnn features,'' \emph{Future Gener. Comput. Syst.}, vol. 125, pp. 820--830, 2021.

\bibitem{Cheng2023MultiscaleSC}
Q.~Cheng, J.~Cheng, Z.~Ren, Q.~Zhang, and J.~Liu, ``Multi-scale spatial–temporal convolutional neural network for skeleton-based action recognition,'' \emph{Pattern Analysis and Applications}, vol.~26, pp. 1303 -- 1315, 2023.

\bibitem{Zhang2019SemanticsGuidedNN}
P.~Zhang, C.~Lan, W.~Zeng, J.~Xue, and N.~Zheng, ``Semantics-guided neural networks for efficient skeleton-based human action recognition,'' \emph{2020 IEEE/CVF Conference on Computer Vision and Pattern Recognition (CVPR)}, pp. 1109--1118, 2019.

\bibitem{Si2019AnAE}
C.~Si, W.~Chen, W.~Wang, L.~Wang, and T.~Tan, ``An attention enhanced graph convolutional lstm network for skeleton-based action recognition,'' \emph{2019 IEEE/CVF Conference on Computer Vision and Pattern Recognition (CVPR)}, pp. 1227--1236, 2019.

\bibitem{Cheng2020SkeletonBasedAR}
K.~Cheng, Y.~Zhang, X.~He, W.~Chen, J.~Cheng, and H.~Lu, ``Skeleton-based action recognition with shift graph convolutional network,'' \emph{2020 IEEE/CVF Conference on Computer Vision and Pattern Recognition (CVPR)}, pp. 180--189, 2020.

\bibitem{Cheng2020DecouplingGW}
K.~Cheng, Y.~Zhang, C.~Cao, L.~Shi, J.~Cheng, and H.~Lu, ``Decoupling gcn with dropgraph module for skeleton-based action recognition,'' in \emph{European Conference on Computer Vision}, 2020.

\bibitem{Chi2022InfoGCNRL}
H.~gun Chi, M.~H. Ha, S.~geun Chi, S.~W. Lee, Q.-X. Huang, and K.~Ramani, ``Infogcn: Representation learning for human skeleton-based action recognition,'' \emph{2022 IEEE/CVF Conference on Computer Vision and Pattern Recognition (CVPR)}, pp. 20\,154--20\,164, 2022.

\end{thebibliography}

\end{document}